\documentclass[journal]{IEEEtai}

\usepackage{amsmath,amsfonts}
\usepackage{algorithmic}
\usepackage{algorithm}
\usepackage{array}
\usepackage[caption=false,font=normalsize,labelfont=sf,textfont=sf]{subfig}
\usepackage{textcomp}
\usepackage{stfloats}
\usepackage{url}
\usepackage{verbatim}
\usepackage{graphicx}
\usepackage{cite}

\usepackage{algorithm}
\usepackage{algorithmic}
\usepackage{multirow}
\usepackage{booktabs}
\usepackage{float}
\usepackage{makecell}
\usepackage{makecell}
\usepackage{CJKutf8}
\usepackage{color}
\usepackage[colorlinks,urlcolor=red,linkcolor=red,citecolor=green]{hyperref}

\begin{document}
\begin{CJK}{UTF8}{gbsn}

\title{{An Iterative Optimizing Framework for Radiology Report Summarization with ChatGPT}}

\author{Chong Ma, Zihao Wu, Jiaqi Wang, Shaochen Xu, Yaonai Wei, Fang Zeng, Zhengliang Liu, Xi Jiang, Lei Guo,\\ Xiaoyan Cai, Shu Zhang, Tuo Zhang, Dajiang Zhu, Dinggang Shen~\IEEEmembership{Fellow, IEEE},\\ Tianming Liu~\IEEEmembership{Senior Member, IEEE}, Xiang Li

% IEEE Publication Technology,~\IEEEmembership{Staff,~IEEE,}
        % <-this % stops a space
%\thanks{This revised manuscript is submitted on December 14, 2023.}
\thanks{C. Ma, Y. Wei, L. Guo, X. Cai, and T. Zhang are with the School of Automation, Northwestern Polytechnical University, Xi'an, 710072, China. (e-mail: \{mc-npu, rean\_wei\}@mail.nwpu.edu.cn, \{lguo, xiaoyanc, tuozhang\}@nwpu.edu.cn).}
\thanks{J. Wang and S. Zhang are with the School of Computer Science, Northwestern Polytechnical University, Xi'an, 710072, China. (e-mail: jiaqi.wang@mail.nwpu.edu.cn, shu.zhang@nwpu.edu.cn).}
\thanks{X. Jiang is with the Clinical Hospital of Chengdu Brain Science Institute, MOE Key Lab for Neuroinformation, School of Life Science and Technology, University of Electronic Science and Technology of China, Chengdu, 611731, China. (e-mail: xijiang@uestc.edu.cn).}
\thanks{D. Shen is with School of Biomedical Engineering, ShanghaiTech University, Shanghai 201210, China, and Department of Research and Development, Shanghai United Imaging Intelligence Co., Ltd., Shanghai 200030, China, and also with Shanghai Clinical Research and Trial Center, Shanghai, 201210, China. (e-mail: Dinggang.Shen@gmail.com).}
\thanks{Z. Wu, S. Xu, Z. Liu, and T. Liu are with the school of computing, University of Georgia, Athens, GA 30602, USA. (e-mail: \{zw63397, sx76699, zl18864, tliu\}@uga.edu).}
\thanks{D. Zhu is with Department of Computer Science and Engineering, The University of Texas at Arlington, Arlington 76019, USA, (e-mail: dajiang.zhu@uta.edu).}
\thanks{F. Zeng and X. Li are with the Department of Radiology, Massachusetts General Hospital, Boston 02114, USA, (e-mail: \{fzeng1, xli60\}@mgh.harvard.edu).}
}

\markboth{Journal of IEEE Transactions on Artificial Intelligence, Vol. 00, No. 0, Month 2020}
{C. Ma \MakeLowercase{\textit{et al.}}: An Iterative Optimizing Framework for Radiology Report Summarization with ChatGPT}

\maketitle

\begin{abstract}
The ``Impression" section of a radiology report is a critical basis for communication between radiologists and other physicians. {Typically written by radiologists, this part is derived from the ``Findings" section, which can be laborious and error-prone. Although deep-learning based models, such as BERT, have achieved promising results in Automatic Impression Generation (AIG),} such models often require substantial amounts of medical data and have poor generalization performance. {Recently,} Large Language Models (LLMs) like ChatGPT have shown strong generalization capabilities and performance, but their performance in specific domains, such as radiology, remains under-investigated and potentially limited. {To address this limitation, we propose ImpressionGPT, leveraging the contextual learning capabilities of LLMs through our dynamic prompt and iterative optimization algorithm to accomplish the AIG task. ImpressionGPT initially employs a small amount of domain-specific data to create a dynamic prompt, extracting contextual semantic information closely related to the test data. Subsequently, the iterative optimization algorithm automatically evaluates the output of LLMs and provides optimization suggestions, continuously refining the output results. The proposed ImpressionGPT model achieves superior performance of AIG task on both MIMIC-CXR and OpenI datasets without requiring additional training data or fine-tuning the LLMs.} This work presents a paradigm for localizing LLMs that can be applied in a wide range of similar application scenarios, bridging the gap between general-purpose LLMs and the specific language processing needs of various domains.
\end{abstract}

\begin{IEEEImpStatement}
With the advent of Artificial General Intelligence (AGI) and Large Language Model (LLM) such as ChatGPT, we envision that a series of medical text data processing methodologies and the corresponding data management solutions can be replaced by LLMs. Thus, in this work, we leveraged the text understanding and summarization capability of ChatGPT for the task of generating the {``Impression”} section of a radiology report, which is a critical basis for communication between radiologists and other physicians. Here we propose ImpressionGPT, which leverages the in-context learning capability of LLMs by constructing dynamic prompts using domain-specific, individualized data via an iterative optimization approach. We envision that the proposed framework can become a paradigm for similar works in the future to adapt general-purpose LLMs to specific domains via an in-context learning approach.  
\end{IEEEImpStatement}


\section{Introduction}

\IEEEPARstart{T}{ext} summarization is the process of compressing a large amount of text data into a shorter summary while maintaining its coherence and informative properties. It has long been a critical and well-studied research area within the field of Natural Language Processing (NLP). As the volume of digital textual information is growing at an extraordinary rate in both the general and medical domains, the need for efficient and accurate text summarization models grows correspondingly. 
In earlier studies, Luhn \cite{Luhn1958} proposed the first automatic summarization algorithm based on statistical methods. Later, a variety of alternative approaches have been proposed, such as rule-based methods \cite{jing2000sentence}, latent semantic analysis \cite{ozsoy2011text}, and graph-based techniques \cite{erkan2004lexrank}. Although traditional methods significantly advanced the research and application of text summarization, they often lack the ability to capture complex semantics and contextual information to generate human-level summarization performance \cite{el2021automatic}.

The introduction of neural networks and deep learning methods, especially the sequence-to-sequence models that employ encoder-decoder architectures for generating summaries \cite{see2017get} conveyed promising results. These approaches enabled the creation of more fluent and contextually relevant summaries compared with rule-based and statistical methods.
Recently, the field of NLP, including text summarization, has experienced drastic changes with the emergence of large-scale, pre-trained foundational models, such as BERT \cite{kenton2019bert} and GPT \cite{radford2019language}. These models are trained on massive volumes of text data, which enables them to learn rich contextual representations and generate human-like languages. Study \cite{liu2019text} has been conducted to demonstrate that fine-tuning these foundational models on text summarization tasks can lead to state-of-the-art performance, outperforming earlier models by a wide margin.

Radiology reports are pivotal in clinical decision-making since they can provide crucial diagnostic and prognostic information to healthcare professionals \cite{cai2021chestxraybert}. The volume of imaging studies and the complexity of radiology data are both growing at an increasing rate, thus raising an urgent need for efficient language processing, including extracting key information from radiology reports. Text summarization can address this challenge by automatically generating concise, informative, and relevant summaries of radiology reports, thus can significantly enhance clinical workflows, reduce the workload of healthcare professionals, and improve patient care~\cite{liu2019text}. With the help of automatic text summarization methods, healthcare professionals can efficiently identify essential information, which leads to faster decision-making, optimized resource allocation, and improved communication among multidisciplinary teams \cite{el2021automatic}.

Compared with general NLP tasks, radiology report summarization has its own unique challenges. It would be difficult for general-purpose NLP models to accurately capture the key information due to the highly specialized and technical nature of the language used. The potential risks associated with misinterpretation of crucial findings, and the significance of maintaining the contextual and relational aspects of such findings further complicate the task~\cite{gehrmann2018comparing}. In response, multiple specialized language processing methods have been developed for medical text summarization, which can be broadly categorized into three groups: traditional, deep learning-based, and {large language model-based. 
Traditional methods such as \cite{jing2000sentence} lay the foundation for text summarization research.}
% Rule-based extraction \cite{jing2000sentence} and keyword-based summarization \cite{HASSANPOUR201629} are both examples of traditional methods that provided a foundation for text summarization research.
However, they often lacked the ability to capture complex semantics and contextual information.
The introduction of deep learning techniques, such as CNNs, demonstrated superior performance in capturing the unique language features and context of radiology reports \cite{gehrmann2018comparing}. However, these techniques often require large volumes of annotated data for training and are limited to specific tasks and/or domains. Subsequently, with the introduction of the Transformer model, its distinctive multi-head global attention mechanism has led to its widespread application in medical image analysis~\cite{ma2023eye,ma2023rectify}.
%,chen2022mask,yu2023core,xiao2023instruction,li2023artificial
Moreover, Transformer-based models, such as BERT \cite{kenton2019bert} and GPT \cite{radford2019language}, brought the emergence of {Large Language Models (LLMs)} and opened new possibilities for radiology text processing. Fine-tuning BERT on domain-specific data, such as chest radiology reports, demonstrating better capability in capturing clinical context and generating high-quality summaries~\cite{cai2021chestxraybert}. Nonetheless, utilizing pre-trained language models such as BERT still needs a significant volume of annotated data for the downstream tasks (e.g., text summarization). 
{Furthermore, training LLMs like ChatGPT~\cite{openaiIntroducingChatGPT} and GPT-4~\cite{openai2023gpt4} still demands a substantial amount of data, even when this data is unannotated. However, in certain specialized domains, the available data is often extremely limited. Utilizing such limited data is insufficient to train an effective large language model. Even with access to an adequate amount of data, training a model with an immense number of parameters necessitates a considerable amount of resources. Hence, the high demands on data quantity and hardware resources pose significant obstacles to the application of large language models in specialized domains.}

% On the other hand, recent LLMs such as ChatGPT~\cite{openaiIntroducingChatGPT} and GPT-4~\cite{openai2023gpt4} adopt the frameworks of in-context learning, enabling fine-tuning of the models without a large amount of annotated data. Thus, identifying the optimized schemes for in-context learning by designing corresponding prompts becomes an important component in developing specific-purposed LLMs \cite{zhao2023brain}. 

\begin{figure}[htb]
\begin{center}
\includegraphics[width=0.45\textwidth]{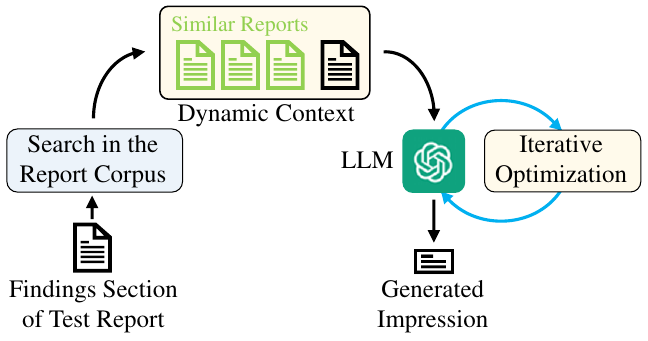}
\end{center}
\caption{Overview of ImpressionGPT. Initially, ImpressionGPT selects similar reports from the corpus based on the {findings section of test report.} Subsequently, a dynamic context is created and fed into ChatGPT. An iterative optimization algorithm is then employed to fine-tune the generated response in an interactive way. Finally, an optimized impression that conforms to the {findings section of test report} is produced.} 
\label{small_pipeline}
\end{figure}

{It is noteworthy that models such as ChatGPT and GPT-4 possess strong in-context learning abilities, allowing them to extract useful semantic information based on provided prompts~\cite{zhao2023brain}. Coupled with their inherent text generation capabilities, even in scenarios with limited training data, these large language models can adapt and generate domain-specific answers through the optimization of prompts.} In this study, we utilized ChatGPT and optimized its generated result for radiology report summarization. An iterative optimization algorithm is designed and implemented via prompt engineering to take advantage of ChatGPT's in-context learning ability while also continuously improving {its response} through interaction. Specifically, as shown in Fig.~\ref{small_pipeline}, we use similarity search techniques to construct a dynamic prompt to include semantically- and clinically-similar existing reports. These similar reports are used as examples to help ChatGPT learn the text descriptions and summarizations of similar imaging manifestations in a dynamic context. We also develop an iterative optimization method to further enhance the performance of the model by performing automated evaluation on the response generated by ChatGPT and integrating the generated text and the evaluation results to compose the iterative prompt. In this way, the generated response from ChatGPT is continually updated for optimized results under pre-defined guidance. We evaluate our ImpressionGPT on two public radiology report datasets: MIMIC-CXR~\cite{johnson2019mimic} and OpenI~\cite{demner2016preparing}. Our experimental results show that ImpressionGPT performs substantially better than the current methods in radiology report summarization, with only a small dataset (5-20 samples) used for {optimizing}. Overall, the main contributions of this work are:
\begin{itemize}
\item In-context learning of {Large Language Model (LLM)} with limited samples is achieved by similarity search. Through the identification of the most similar examples in the corpus, a dynamic prompt is created that encompasses the most useful information for LLM. 
\item {An iterative optimization algorithm is developed with a dynamic prompt scheme to further optimize the generated result.} The iterative prompt provides feedback on the responses generated by LLM and the corresponding evaluation, followed by further instructions to iteratively update the prompts.
\item {A new paradigm for optimizing LLMs' generation using domain-specific data.} The proposed framework can be applied to any scenarios involving the development of domain-specific models from an existing LLM in an effective and resource-efficient approach. And the corresponding code is available on GitHub$\footnote{https://github.com/MoMarky/ImpressionGPT}$.
\end{itemize}

\section{Background and Related Works}

\subsection{Text Summarization in Natural Language Processing}

Text summarization is to extract, summarize or refine the key information of the original text to obtain the main content or general meaning of the the original text. There are two major categories of text summarization: extractive summarization and abstractive summarization. Among them, extractive summarization~\cite{erkan2004lexrank,jing2000sentence,liu2019fine,ozsoy2011text} is to take one or more sentences from a text or text set to construct a summary. An advantageous aspect of this approach lies in its simplicity, and its results exhibit a low tendency towards deviations from the essential message conveyed in the text. Before the emergence of artificial neural network, the practice of text summarization predominantly relied on the technique of extractive summarization. For example, Hongyan~\cite{jing2000sentence} developed filtering rules based on prior knowledge to remove unimportant parts of the text to obtain a summary. LexRanK~\cite{erkan2004lexrank} represented the words and sentences in a graph and search the text with the highest similarity in the graph. \cite{ozsoy2011text} used algebraic statistics to extract latent semantic information from text and generates text summaries based on latent semantic similarities. After the emergence of artificial neural networks and deep learning, methods such as BertSUM~\cite{liu2019fine} and TransEXT~\cite{liu2019text} were training based on the BERT~\cite{kenton2019bert} model. However, extractive summarization suffers from incoherent generation of summaries, uncontrollable length, and the quality of the results is severely dependent on the original text. 

For the task of abstractive summarization~\cite{lin2018global,liu2019fine,liu2019text,see2017get,song2020controlling}, there is no issue as mentioned previously. Abstractive summarization task is an end-to-end generative task, and it is necessary to understand the meaning of the original text and generate a new summarization. Compared with extractive summarization, abstractive summarization is more challenging, but it is also more in line with the daily writing habits of human beings, so it has gradually become a significant research focus in the field of text summarization since the introduction of artificial neural networks and deep learning. For example, PGN (LSTM)~\cite{see2017get} used a pointer-generator network to copy words from the original text and also retains the ability to generate new words, thus improving the seq2seq+attention model architecture. Similarly, CGU~\cite{lin2018global} proposed a global encoding framework for summary generation, which uses a combination of CNN and self-attention to filter the global encoding of text, solving the alignment problem between source text and target summary. With the emergence of BERT~\cite{kenton2019bert}, a milestone model within the NLP field, the previous training method was changed to use the strategy of pre-training + fine-tuning. 
% Among others, BART~\cite{lewis2020bart} explored the use of pre-trained models to achieve advanced performance on text summarization task. 
TransABS~\cite{liu2019text} accomplished generative summarization based on the BertSUM~\cite{liu2019fine} used a two-stage fine-tuning method and achieved optimality on three datasets. CAVC~\cite{song2020controlling} used a Mask Language Modeling (MLM) strategy based on the BERT model to further improve the performance. 
% There are also some approaches that try to use the pre-training model in the medical-related text processing, such as, \cite{park2020continual} helped the scientific community to understand the rapidly flowing COVID-19 literature array based on the BERT model. \cite{kieuvongngam2020automatic} used the BERT and GPT~\cite{chang2021jointly} to generate text summaries of COVID-19 medical research articles. 
 
\subsection{Radiology Report Summarization}
With the development of text summarization in NLP, text processing related to the medical field is also gaining attention. In the standard radiology report, the impression section is a summary of the entire report description. Therefore, {Automatic Impression Generation} (AIG) has become the focus of NLP research in the medical field~\cite{karn2022differentiable,lee2020biobert,rezayi2022clinicalradiobert,hu2021word,cai2021chestxraybert,hu2022graph}. Earlier studies have focused on the use of seq2seq methods. For example, \cite{karn2022differentiable} trained a bi-directional LSTM based model as a summary extractor using the method of Multi-agent Reinforcement Learning. 
% \cite{gharebagh2020attend} extracted salient clinical ontology terms from the study results and then incorporated them into a summarizer using a separate encoder. 
% \cite{zhang2018learning} first employed a bi-directional LSTM for sequence-to-sequence generation and found that 30\% of the generated radiology summaries had factual errors.

After the emergence of BERT~\cite{kenton2019bert}, BioBERT~\cite{lee2020biobert} pre-trained BERT using large-scale biomedical corpora, surpassing previous methods in a variety of medical-related downstream tasks such as named entity recognition, relationship extraction, and question answering.
% ~\cite{liao2023mask,rezayi2022clinicalradiobert}. 
This work explored the path of using pre-trained language models within the biomedical domain. 
Similar, ClinicalRadioBERT~\cite{rezayi2022clinicalradiobert} pre-trained a BERT model and proposed a knowledge-infused few-shot learning (KI-FSL) approach that leverages domain knowledge for understanding radiotherapy clinical notes.
% Similar, Clinical BERT~\cite{huang2019clinicalbert} pre-trained BERT using clinical notes and fine-tuned it for the task of predicting hospital readmission. 
ChestXrayBERT~\cite{cai2021chestxraybert} pre-trained BERT using a radiology-related corpus and combined it as an encoder with a Transformer decoder to perform the diagnostic report summarization task.
And WGSUM~\cite{hu2021word} constructed a word graph from the findings section of radiology report by identifying the salient words and their relations, and proposed a graph-based model WGSUM to generate impressions with the help of the word graph. \cite{hu2022graph} utilized a graph encoder to encode the word graph during pre-training to enhance the text extraction ability of the model, and introduced contrast learning to reduce the distance between keywords. Its results on AIG outperform the previous methods. 
All the above methods have achieved good results in the medical text domain based on the pre-trained language model, but still have the problem of poor generalization due to the low complexity of the model.

\subsection{Large Language Model}
With the advent of BERT~\cite{kenton2019bert} model based on Transformer architecture, an increasing number of studies related to {Natural Language Processing} (NLP) have incorporated pre-training + fine-tuning methodologies. The approach that first pre-training on a large amount of unlabeled data and then fine-tuning on a small portion of labeled data has proven to achieve more outstanding results. For example, before BERT, GPT-1~\cite{liu2018generating} with 117 million parameters has been initially trained using self-supervised pre-training $+$ supervised fine-tuning. It directly used the Transformer decoder to achieve excellent results on natural language inference and question-and-answer tasks. Later, Google proposed the landmark model BERT, which introduced Transformer encoder and further improved the performance by using Mask Language Modeling and Next Sentence Prediction methods in the pre-training stage, where the number of parameters in BERT-Large has reached 340 million. Four months after the release of BERT, GPT-2~\cite{radford2019language} was introduced, which further extended the model parameters and training data set based on GPT-1, with Extra Large of GPT-2 model reaching 1.5 billion parameters. In addition, the researchers~\cite{radford2019language} found that with the expanded training dataset, outstanding results of large language model could be achieved in downstream tasks without using fine-tuning. GPT-3~\cite{brown2020language} further expanded the data size and parameter size based on GPT-2, and the maximum parameter reached 175 billion, and its performance on downstream tasks was significantly improved. And they first proposed a training paradigm of unsupervised pre-training $+$ few-shot prompt.
% Recently, PaLM~\cite{chowdhery2022palm} used Pathways~\cite{barham2022pathways} to train a model with 540 billion parameters, and further refreshed the results using zero-shot, one-shot and few-shot in downstream tasks.

In comparison to small {Pre-trained Language Models} (PLMs), LLMs possess superior generalization capability. They can accurately learn potential features of input text and perform effectively across different downstream tasks, even without fine-tuning.
% ~\cite{liu2023tailoring,,liu2023radonc,dai2023ad,holmes2023evaluating}. 
One prominent foundational model of a large language model is ChatGPT~\cite{openaiIntroducingChatGPT}, based on the GPT-3.5 model, which employs training data in conversation mode to facilitate user-friendly human-machine interaction. ChatGPT has been widely integrated into various applications such as education and healthcare, and performs well in tasks such as text classification, data expansion, summarization, and other natural language processing~\cite{liu2023tailoring,shi2023mededit,dai2023auggpt,liu2023summary,liu2023deid,tang2023policygpt}.
% ~\cite{dai2023auggpt,tang2023policygpt,liu2023deid,liu2023summary}.
Although ChatGPT performs well in most tasks, its performance in specialized domains is still unsatisfactory. Hence, we propose ImpressionGPT, an iterative optimization algorithm that enables ChatGPT to achieve excellent performance on the radiology report summarization task.

\subsection{Prompt Engineering}
Prompt engineering is a burgeoning field that has garnered significant attention in recent years due to its potential to enhance the performance of {LLMs}. The fundamental idea behind prompt engineering is to utilize prompts as a means to program LLMs, which is a vital skill required for effective communication with these models~\cite{wang2023prompt}, such as ChatGPT. Recent research has demonstrated that designing prompts to guide the model toward relevant aspects of input can lead to more precise and consistent outputs. This is particularly crucial in applications such as language translation and text summarization, where the quality of the output is paramount.

In general, prompt engineering is a new paradigm in the field of natural language processing, and although still in its early stages, has provided valuable insights into effective prompt patterns~\cite{liu2023pre}. These patterns provide a wealth of inspiration, highlighting the importance of designing prompts to provide value beyond simple text or code generation.
% ~\cite{xiao2023instruction,dai2023samaug,wang2023prompt,wang2023review}. 
However, crafting prompts that are suitable for the model can be a delicate process. Even a minor variation in prompts could significantly impact the model's performance. Therefore, finding the most suitable prompts remains an important challenge. Typically, there are two main types of prompts: manual prompts and automated template prompts.

\subsubsection{Manual Prompt}
Manual prompts are designed manually to guide LLMs towards specific inputs. These prompts provide the model with explicit information about what type of data to focus on and how to approach the task at hand~\cite{liu2023pre}. Manual prompts are particularly useful when the input data is well-defined and the output needs to adhere to a specific structure or format~\cite{brown2020language}. For example, in the medical field where interpretability is crucial, manually created prompts are often employed to guide the model toward focusing on specific aspects of the input data. In the area of medical text security, for instance, manual prompts were utilized to guide the model toward identifying and removing private information in medical texts, effectively solving the ethical issues associated with medical data~\cite{liu2023deid}. Overall, manual prompts play a vital role in improving model performance in various domains by providing the model with a more structured and focused approach to the task.

\subsubsection{Automated Template Prompt}
While manual prompts are a powerful tool for addressing many issues, they do have certain limitations. For example, creating prompts requires time and expertise, and even minor modifications to prompts can result in significant changes in model predictions, particularly for complex tasks where providing effective manual prompts is challenging~\cite{jiang2020can}. To address these issues, researchers have proposed various methods for automating the process of prompt design, different types of prompts can assist language models in performing specific tasks more effectively. Prompt mining involves extracting relevant prompts from a given dataset~\cite{jiang2020can}, while prompt paraphrasing improves model performance and robustness by increasing prompt diversity. Gradient-based search~\cite{shin2020autoprompt} helps identify optimal prompts in a model's parameter space, and prompt generation~\cite{ben2021pada} can create new prompts using techniques such as generative models. These discrete prompts are typically automatically searched for in a discrete space of prompt templates, often corresponding to natural language phrases. Other types of prompts, such as continuous prompts~\cite{tsimpoukelli2021multimodal} that construct prompts in a model's embedding space and static prompts that create fixed prompt templates for input can also aid in task performance.
% , and dynamic prompts that generate custom templates for each input, ~\cite{liu2022declaration}.

These different prompt types can be used independently or in combination to help language models perform various tasks, including natural language understanding, generation, machine translation, and question-answering.

\section{Method}

\begin{figure*}[htb]
\begin{center}
\includegraphics[width=0.95\textwidth]{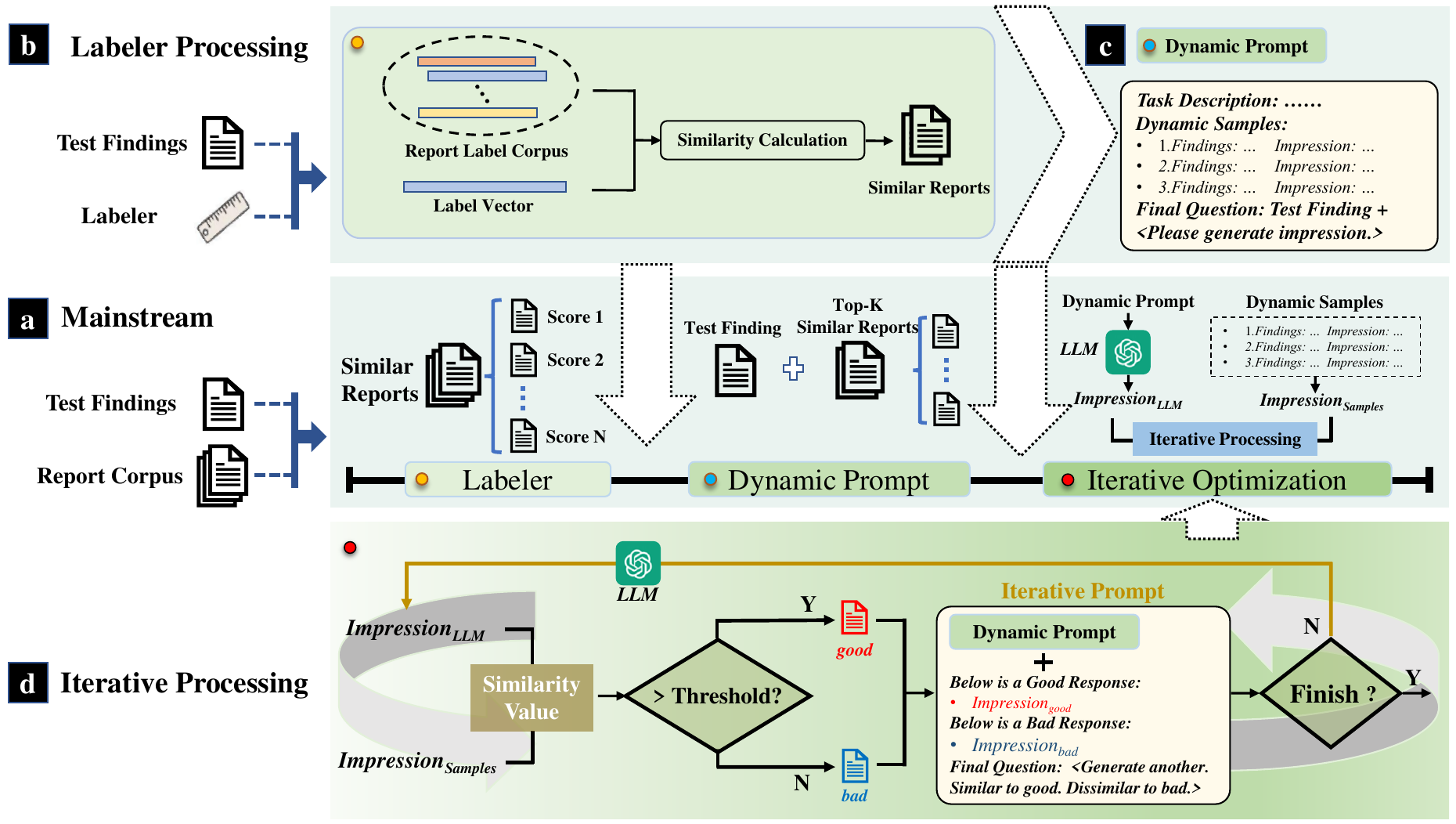}
\end{center}
\caption{The pipeline of our ImpressionGPT. Part $a$ in the middle is the mainstream of our method. We first use a labeler to categorize the diseases of test report and obtain the similar reports in the corpus (part $b$), and then construct a dynamic prompt in part $c$. Part $d$ accomplishes the iterative optimization of {LLM's generated result} through interaction with positive (\textcolor{red}{$good$}) and negative (\textcolor{blue}{$bad$}) responses.
} 
\label{pipeline}
\end{figure*}

In this section, we first illustrate the pipeline of our ImpressionGPT. Then we elaborate on the dynamic prompt generation in Sec.~\ref{dynamic_prompt_generation} and the iterative optimization in Sec.~\ref{iterative_opti}.
% In this work, we employed dynamic prompt and iterative optimization to enhance the adaptation of ChatGPT to radiology report summarization. 
{Fig.~\ref{pipeline} shows the pipeline of our ImpressionGPT. Firstly, as shown in the first step of mainstream (part $a$ in Fig.~\ref{pipeline}),} we use a labeler to categorize the {``Findings"} section of the report and extract disease labels. Then, based on the disease category, we search for similar reports in the existing diagnostic report corpus, as shown in part {$b$} of Fig.~\ref{pipeline}. And we designed a dynamic prompt (shown in part {$c$} of Fig.~\ref{pipeline}) to construct a context environment with similar diagnostic reports, so that ChatGPT can learn to summarize diagnostic reports related to the current disease. We refer to this as {``dynamic context"}. Based on the dynamic context, we utilized an iterative optimization method, as shown in the part {$d$} of Fig.~\ref{pipeline}, to {optimize} the response of ChatGPT. During the iterative optimization process, we compare the generated {``Impression"} from ChatGPT with examples in dynamic prompt to obtain good and bad responses. {These evaluated responses are inserted to the iterative prompt with further guidance to ensure that the next response is closer to the former good response while avoiding the former bad response.} Overall, our method requires a small number of examples, facilitating ChatGPT's acquisition of excellent domain-specific processing capabilities. More details of dynamic prompt generation and iterative optimization can be found in the following subsections.

\subsection{Dynamic Prompt Generation}
\label{dynamic_prompt_generation}
In previous manual-designed prompts, fixed-form prompts were frequently employed for simple tasks that were easily generalized, such as translation, Q\&A, and style rewriting. However, these fixed-form prompts were found to be insufficient in providing prior knowledge for more intricate tasks and datasets that are peculiar to specific domains, like processing medical diagnosis reports, resulting in poor performance of ChatGPT. Consequently, we propose a hypothesis which suggests that constructing dynamic prompts by utilizing similar examples from relevant domain-specific corpora can enhance the model's comprehension and perception. {In this work, dynamic prompt generation primarily comprises two main components: similarity search and prompt design. Below is a detailed introduction to these two components.}

\subsubsection{Similarity Search}
\label{similarity_search}
{The purpose of similarity search is to find diagnostic reports within the corpus that are similar to the Test Findings. The process for similarity search mainly encompasses two phases.} Initially, a disease classifier is employed to extract the disease categories appearing in the input radiology report. {Subsequently, relying on these categories, a similarity calculation is conducted on the report corpus to obtain examples similar to the input radiology report.}

\begin{figure*}[htb]
\begin{center}
\includegraphics[width=0.7\textwidth]{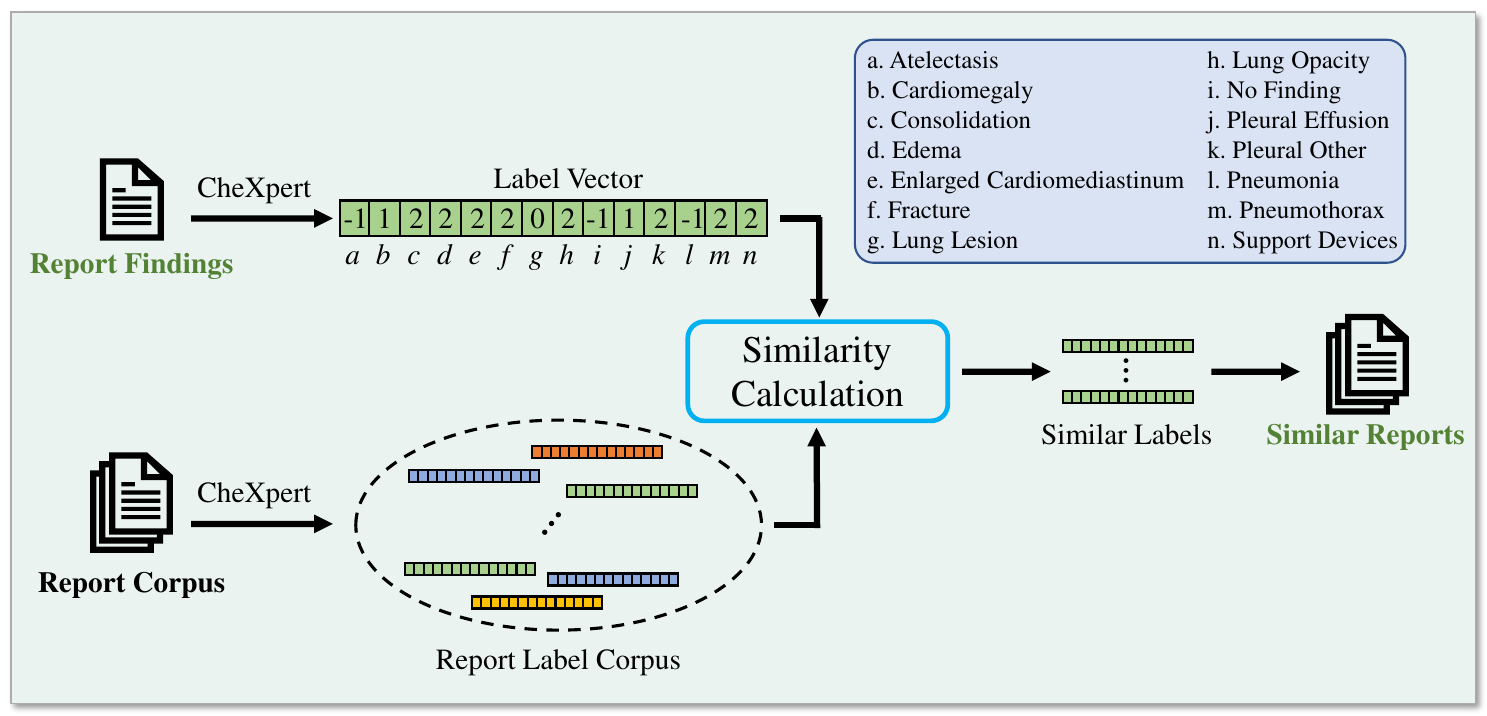}
\end{center}
\caption{Details of similarity search process. CheXpert is initially employed to obtain the label vector for the radiology reports present in the corpus. Then the similarity between the label vector of the test report and the label corpus is calculated, allowing for the identification of the similar radiology reports.} 
\label{similarity_search_fig}
\end{figure*}

{In previous studies, corpus search typically involve two approaches: text-based~\cite{karthiga2022similarity} and feature-based~\cite{qurashi2020document}. However,} text-based methods can be time-consuming and significantly increase search time, particularly with large corpora. Feature-based methods require feature extraction and storage of each sample, making it more space-demanding. {Therefore, in this study,} we utilize diagnostic report labels for similarity search, requiring only prior identification and local label storage of the report sample. Our approach substantially reduces time and space costs. Moreover, using tag values to calculate the similarity enables the identification of samples with similar diseases, which provides domain-specific and even sample-specific knowledge that can be leveraged by ChatGPT's contextual processing capabilities. {Specifically,} we use CheXpert labeler~\cite{irvin2019chexpert} as the disease classifier, which is a rule-based labeler to extract observations from the free text radiology reports. The observations contain 14 classes based on the prevalence in the chest radiology reports. As shown in Fig.~\ref{similarity_search_fig}, each observation is represented by a letter from {`$a$'} to {`$n$'}, and each contains four categories. 
The categories correspond to the clearly presence {(`$1$')} or absence {(`$0$')} of the observation in the radiology report, the presence of uncertainty or ambiguous description {(`$-1$')}, and the unmentioned observation ($blank$), which is replaced by the number {`$2$'} to facilitate similarity calculation. The labels of each radiology report in the corpus are extracted and saved as a one-dimensional vector of scale $1 \times 14$. The label vectors of test reports are then compared to those in the corpus using Euclidean distance, and the radiology reports that are closest to each other are selected to generate dynamic prompts. This method ensures reliable and accurate disease classification of radiology reports. In this work, we directly use the split training set as a corpus and extract and preserve the corresponding disease labels for MIMIC-CXR~\cite{johnson2019mimic} and OpenI~\cite{demner2016preparing} datasets. {More details about dataset processing can be found at Section~\ref{dataset_metric}.}

% For the MIMIC-CXR~\cite{johnson2019mimic} dataset, we use the officially split training set as a corpus and extract and preserve the corresponding disease labels. To minimize the time required for similarity computation in cases of particularly large corpora like MIMIC-CXR dataset, we employ a systematic random sampling technique to generate a subset of the original corpus. Specifically, 10,000 random radiology reports of the 122,014 training data from MIMIC-CXR are selected for the current similarity computation. On the other hand, since the OpenI~\cite{demner2016preparing} dataset is relatively small in volume, we directly use all available samples in its training set as the corpus.

\subsubsection{{Prompt Design}}
\label{dynamic_prompt}

\begin{figure*}[ht]
\begin{center}
\includegraphics[width=0.65\textwidth]{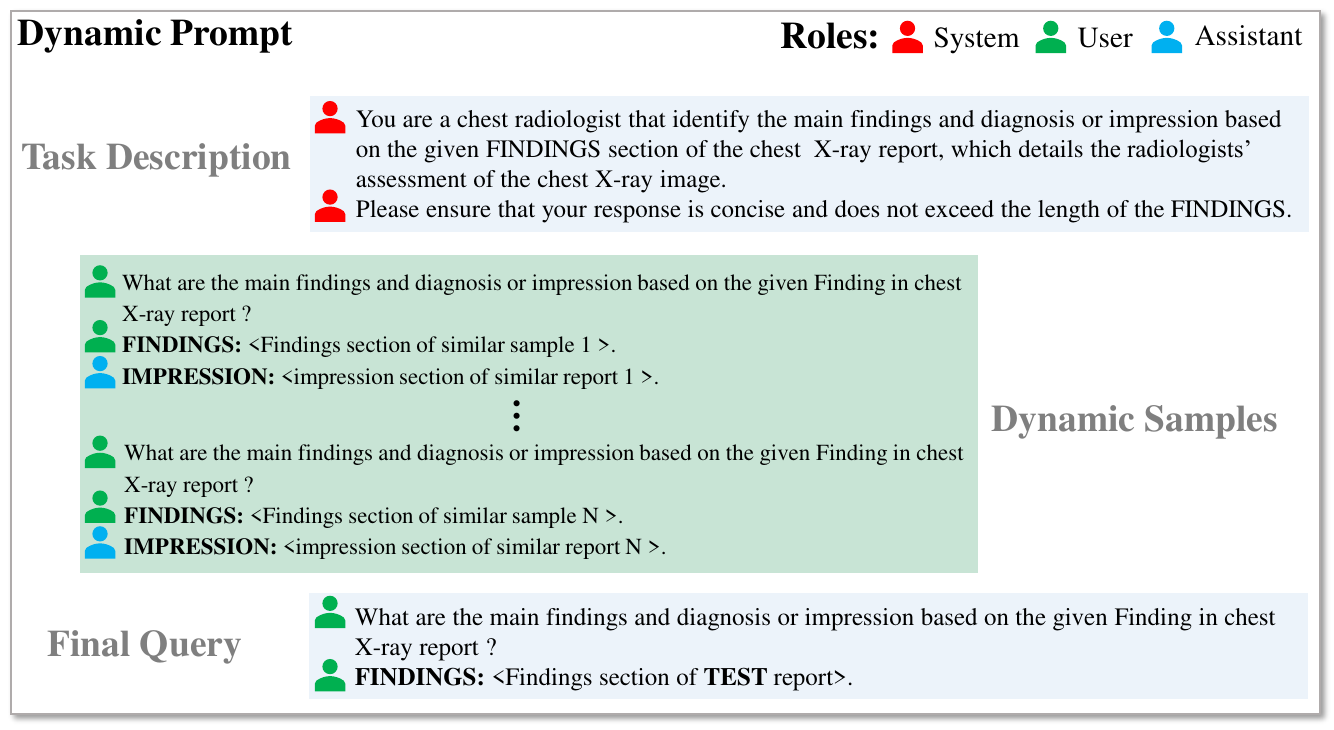}
\end{center}
\caption{Details of dynamic prompt. The dynamic prompt is composed of task description, dynamic samples, and final query. Each component contains message of multiple roles. The flags in front of each sentence represent the role of the current message, \textcolor{red}{red} for System, \textcolor{green}{green} for User, and {blue} for Assistant.} 
\label{dynamic_prompt_fig}
\end{figure*}

{Prior to introducing the design of our dynamic prompt,} it is necessary to provide an overview of the input format when utilizing the ChatGPT API. The message that is fed into ChatGPT is accessed through the API in the form of a string, which is categorized into three distinct roles: $System$, $User$, and $Assistant$, as illustrated in Fig.~\ref{dynamic_prompt_fig}. These roles are represented by the colors red, green, and blue, respectively. The $System$ message initiates the conversation and provides information regarding the task while constraining the behavior of the $Assistant$. The $User$ message instructs the $Assistant$ and serves as the input provided by the user. Lastly, the $Assistant$ component represents the response that is generated by the model. Notably, as the $Assistant$ message can be artificially set, it enables the implementation of our dynamic prompt and facilitates iterative optimization. 

In this work, we employ the prefix form for designing the dynamic prompt. The dynamic prompt consists of three modules: task description, dynamic samples, and final query, as illustrated in Fig.~\ref{dynamic_prompt_fig}. The task description module specifies the role of the ChatGPT as a chest radiologist and provides a brief overview of the radiology report summary task along with a simple rule that serves as the foundation for the entire prompt. Subsequently, the dynamic prompt integrates similar reports obtained in Sec.~\ref{similarity_search}. The dynamic samples module, depicted in the central part of Fig.~\ref{dynamic_prompt_fig}, utilizes a question-and-answer format to provide the prompt in each sample. Specifically, the question part consists of a pre-defined question sentence and the {``Findings"} section of dynamic sample and is treated as the message of the $User$ role. Then the {``Impression"} section of the dynamic sample is treated as the following message of the $Assistant$ role. At the end of our dynamic prompt, the same pre-defined question is used as in the previous samples, and the {``Findings"} section of the test report is inserted. Overall, the resulting dynamic prompt uses a question-answering approach and provides multiple examples that have similar content to the target test sample, thus creating a data-specific dynamic context.

\subsection{{Iterative Optimization of Response}}
\label{iterative_opti}

% To further improve ChatGPT's ability to summarize radiology report, we used an iterative optimization algorithm that first evaluates the response generated by ChatGPT using the dynamic prompt designed in Sec.~\ref{dynamic_prompt} to identify good and bad responses, after which these responses are placed in the dynamic prompt with more detailed instructions again into ChatGPT. With our additional instructions, ChatGPT iteratively optimizes the response closer to the evaluated good ones and avoids the bad ones. The detailed algorithm is shown in Algorithm~\ref{algorithm_1}. 

In Sec.~\ref{dynamic_prompt_generation}, we constructed a dynamic context within the prompt in order to facilitate the model learning relevant prior knowledge from semantically similar examples. However, the use of this fixed form of prompt on ChatGPT produced a one-off effect, as we cannot guarantee that the generated response is appropriate. {Therefore, based on the dynamic prompt, we utilized an iterative optimization operation to further optimize the generated result of ChatGPT. The specific optimization process is presented in Algorithm~\ref{algorithm_1}.} 
% The first step is to input the dynamic prompt designed in Sec.~\ref{dynamic_prompt_generation} and $N_S$ similar radiology reports we selected. 

\begin{algorithm}
    \caption{Iterative Optimization Algorithm}
    \label{algorithm_1}
    \begin{algorithmic}[1]
        \STATE \textbf{Input:} Dynamic prompt with $N_s$ similar reports
        \STATE \textbf{Initialize:} $I$=Iteration times, $Thre_S$=Threshold of evaluation score, $iter$=0
        \STATE \textbf{Definition:} $GPT$ means ChatGPT, $Prompt_{Dy}$ means our dynamic prompt, $Prompt_{Iter}$ means our iterative prompt, $L$ means evaluate function, $M_{good}$ and $M_{bad}$ represent prompt merging with good and bad response
        \WHILE {$iter < I$}
            \IF{$iter=0$}
                \STATE{$response = GPT(Prompt_{Dy})$}
            \ELSE
                \STATE{$response = GPT(Prompt_{Iter})$}
            \ENDIF
            \STATE{$score = \frac{1}{N_s} \sum\limits_{i=0}^{N_s} L(response, Impression_i) $}
            \IF{$score > Thre_S$}
                \STATE{$Prompt_{Iter} = M_{good}(Prompt_{Dy}, response)$}
            \ELSE
                \STATE{$Prompt_{Iter} = M_{bad}(Prompt_{Dy}, response)$}
            \ENDIF  
            \STATE{$iter++$}
        \ENDWHILE
    \end{algorithmic}
\end{algorithm}

% First, the input of the algorithm is the dynamic prompt designed in Sec.~\ref{dynamic_prompt_generation} and the $N$ similar radiology reports we selected. In the initial iteration, the dynamic prompt is directly inputted into ChatGPT. After receiving the initial response, we evaluate it with a defined threshold. As described in lines 10-14 of Algorithm~\ref{algorithm_1}, if the evaluated score is higher than the previously defined threshold, it is considered acceptable and is merged into the prompt to generate the iterative prompt. Conversely, if the score is lower than the threshold, the response is considered unsatisfactory and also merged into the iterative prompt. The purpose of the iterative prompt is to enable ChatGPT to optimize its response during the iterative process, so that the responses are more similar to those considered good and avoid the ones that are considered bad. Through this method, ChatGPT becomes iterative and self-optimizing. A detailed description of the evaluation method and iterative prompt design is described below. 

{The first input of the algorithm is the dynamic prompt ($Prompt_{Dy}$) constructed in Sec.~\ref{dynamic_prompt_generation}. In line 6 of Algorithm~\ref{algorithm_1}, we receive a initial $response$ from ChatGPT based on our dynamic prompt. Then, we evaluate it with other impression section of similar report and calculate an evaluation $score$ at line 10. In lines 11-15, we compare above $score$ with a predefined threshold ($Thre_S$) and construct our iterative prompt. Finally we feed this iterative prompt into ChatGPT to generate an optimized response ($response$ in line 8) and repeat the above procedures.} The purpose of the iterative prompt is to enable ChatGPT to optimize its response during the iterative process, so that the responses are more similar to those considered good and avoid the ones that are considered bad. Through this method, ChatGPT becomes iterative and self-optimizing. A detailed description of the evaluation method and iterative prompt design is described below.

\subsubsection{Response Evaluation}
\label{response_evaluation}

During the process of iterative optimization, it is crucial to assess the quality of a response. In this study, we employ the Rouge-1 score, a prevalent metric for evaluating summarization models. The similarity between the generated results and ground-truth is measured on a word-by-word basis by the Rouge-1 score, thus enabling the evaluation of the outputs at a more refined scale. The dynamic prompt incorporates $N$ radiology reports having the highest similarity in the corpus, with the impression section deemed of high reference value for the ChatGPT response evaluation. As depicted in line 10 of Algorithm~\ref{algorithm_1}, we calculate the Rouge-1 score of ChatGPT's response with the impression section ($Impression_i$) of each similar report in the dynamic prompt, and average it to get the final evaluation result ($score$ in line 10) . Subsequently, we derive ``good response" and ``bad response" based on the pre-defined threshold and merge the outcomes into the iterative prompt ($Prompt_{Iter}$). Specifically, if the evaluated score is higher than the previously defined threshold ($score > Thre_S$), it is considered as a good response. Conversely, if the score is lower than the threshold, the response is considered as a bad response. We update the iterative prompt by integrating the feedback through our merging function ($M_{good}$ and $M_{bad}$). After constructing the iterative prompt, we feed it again into ChatGPT. Note that, this process is repeated until the maximum number of iterations ($iter = I$). {Details of iterative prompt design are described in the following subsection.}

\subsubsection{{Prompt Design}}
\label{iterative_prompt_design}

\begin{figure*}[htb]
\begin{center}
\includegraphics[width=0.65\textwidth]{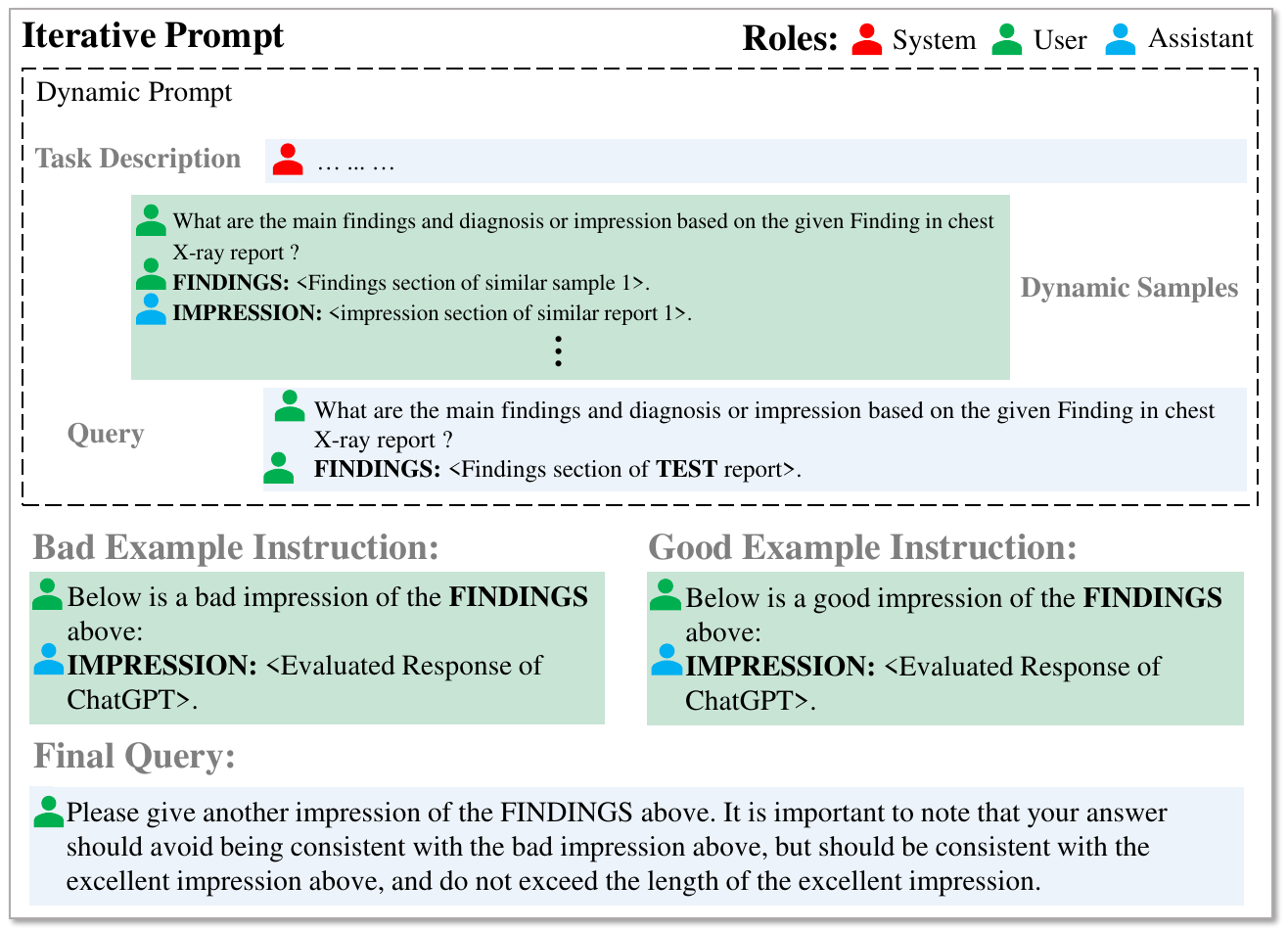}
\end{center}
\caption{Details of iterative prompt. We construct our iterative prompt by including the bad and good example instructions along with the final query, in addition to applying a {``Instruction $+$ Response"} approach in the example instruction for optimizing the response.} 
\label{iterative_prompt_fig}
\end{figure*}

The subsequent and crucial stage after identifying the positive and negative feedback is to utilize the evaluation results to instruct ChatGPT in generating enhanced responses. As shown in Fig.~\ref{iterative_prompt_fig}, the initial part of the iterative prompt is the dynamic prompt described in Sec.~\ref{dynamic_prompt}, which delivers the fundamental context for subsequent enhancement. For the responses evaluated in Sec.~\ref{response_evaluation}, a pair of $User$ and $Assistant$ roles are included in the iterative prompt. The $User$ initiates the message with {``Below is an excellent impression of the FINDINGS above"} for the positive feedback and {``Below is a negative impression of the FINDINGS above"} for the negative feedback. The response produced by ChatGPT is positioned after the corresponding $User$ message, adopting the {``Instruction $+$ Response"} form to enable ChatGPT to learn the relevant content from the good and bad samples. As shown in the bottom {``Final Query"} in Fig.~\ref{iterative_prompt_fig}, the final query that concludes the iterative prompt provides specific optimization rules to regenerate a response that is consistent with the good response and avoids the bad response. Additionally, a length limit instruction is included to avoid ChatGPT from generating overly verbose responses. {Note that after extensive experimental validation, we finally insert one good and $n$ bad responses in the iterative prompt. The number of bad examples ($n$) is indefinite and can be appended all the time as long as the input limit of ChatGPT is not exceeded. We conducted ablation experiments and discussion on the setting of the number of good and bad responses in Section I of supplementary materials.} In the process of each iteration, good response or bad response will be updated, which enables our prompt to optimize ChatGPT's responses in time.

To summarize, our approach involves utilizing a dynamic prompt to establish a contextual understanding that is highly relevant to the semantics of the given test case. This context is then fed into ChatGPT to obtain an initial response, which is further evaluated and incorporated into an iterative prompt. The iterative prompt is used to elicit subsequent responses that are specific to the task domain, thereby enabling self-iterative updates of {ChatGPT's generated result} with limited examples. 
Our experimental results demonstrate that ImpressionGPT delivers superior results in the summarization of radiology reports, as elaborated in Sec.~\ref{compare_to_prominent_methods}. 
% We also study the optimal choice of all parameters in Algorithm~\ref{algorithm_1} in Section I of supplementary materials.

\section{Experiment and Analysis}

\subsection{Dataset and Evaluation Metrics}
\label{dataset_metric}
We evaluated our ImpressionGPT on two public available chest X-ray datasets: MIMIC-CXR~\cite{johnson2019mimic} and OpenI~\cite{demner2016preparing}. MIMIC-CXR dataset contains of 227,835 radiology reports, which are amalgamated into our corpus after applying the official split that includes both the training and validation sets. The test set is reserved solely for evaluating the effectiveness of our ImpressionGPT. In line with the objective of medical report summarization, we excluded ineligible reports by implementing the following criteria: (1) removal of incomplete reports without finding or impression sections, (2) removal of reports whose finding section contained less than 10 words, and (3) removal of reports whose impression section contained less than 2 words. Consequently, we filtered out 122,014 reports to construct our corpus and included 1,606 test reports. 
OpenI dataset contains 3,955 reports, and we use the same inclusion criteria as above and finally obtained 3419 available reports. Since the official split is not provided, we follow~\cite{hu2021word} to randomly divided the dataset into train/test set by 2400:576. Similar to the MIMIC-CXR dataset, we use the training set to build our corpus, while the test set is reserved for evaluating the model.

\setlength{\tabcolsep}{4pt}
\begin{table}[htb]
\small
\centering
\caption{The statistics of the two benchmark datasets with official split for MIMIC-CXR and random split for OpenI after pre-processing. We report the averaged sentence-based length (AVG. SF, AVG. SI), the averaged word-based length (AVG. WF, AVG. WI) of both {``Findings" and ``Impression"} sections.}
\label{table_stat}
\scalebox{0.9}{\begin{tabular}{lllll}
\noalign{\smallskip}
\hline
\noalign{\smallskip}
\multirow{2}{*}{Type} &\multicolumn{2}{c}{MIMIC-CXR~\cite{johnson2019mimic}} &\multicolumn{2}{c}{OpenI~\cite{demner2016preparing}}\\  \cmidrule(r){2-3} \cmidrule(r){4-5}
 &Train &Test &Train  &Test\\
\toprule
Report Num
&122,014    &1,606   &2,400    &576  \\
\hline
\noalign{\smallskip}
AVG. WF 
&55.78    &70.67    &37.89    &37.98        \\
AVG. SF
&6.50    &7.28    &5.75    &5.77       \\
\hline
\noalign{\smallskip}
AVG. WI
&16.98    &21.71    &10.43    &10.61       \\
AVG. SI
&3.02    &3.49    &2.86    &2.82       \\
\bottomrule
\end{tabular}}
\end{table}
\setlength{\tabcolsep}{1.4pt}

Table~\ref{table_stat} shows the statistics of MIMIC-CXR and OpenI datasets after pre-processing. The code to process radiology reports is publicly available on GitHub$\footnote{https://github.com/MoMarky/radiology-report-extraction}$. In our experiments, we employ Rouge metrics to evaluate the generated impression from our ImpressionGPT, and we reported F1 scores for Rouge-1 (R-1), Rouge-2 (R-2), and Rouge-L (R-L) that compare word-level unigram, bi-gram, and longest common sub-sequence overlap with the reference impression of the test report, respectively. {In addition to using Rouge metrics, we incorporate the factual consistency (FC) metric, as introduced in previous works~\cite{hu2021word,hu2022graph}. The FC metric measures similarity between reference impressions and generated impressions by incorporating generic disease-related observations.
% The FC metric involves employing CheXbert~\cite{smit2020chexbert} to identify 14 disease-related observations within both reference impressions and generated impressions.
Precision, recall, and F1 score are subsequently utilized to assess the performance. Given that existing studies only provide FC metrics for the MIMIC-CXR dataset, we also compute the Precision (FC-P), Recall (FC-R), and F1 score (FC-F1) for ImpressionGPT specifically on the MIMIC-CXR dataset.}

\subsection{{Implementation Details}}
{In our framework, dynamic prompt construction and iterative optimization of response are essential components, and therefore, the related parameters are crucial. Specifically, in dynamic prompt construction, we define the number of similar reports as $N_s$, which determines the richness of dynamic semantic information in our prompt. In the iterative optimization of response, we define the evaluation threshold as $T$, the number of good response and bad response after evaluation as $G_d$ and $B_d$, and the number of iterations as $I$. The parameter $T$ determines the quality of ChatGPT's response during the optimization process. $G_d$ and $B_d$ determine the richness of positive and negative feedback information in the iterative prompt. And $I$ determines the degree of optimization of ChatGPT's response in our framework.}

{We conducted detailed ablation experiments on all parameters in supplementary materials and obtained the optimal values of $N_s=15, T=0.7, G_d=1, B_d=n (n>1), I=17$. It should be noted that there is no upper limit set for the value of $B_d$, and the value of $n$ is flexibly adjusted based on the length of the actual input information during the optimization process. Detailed experimental analysis can be found in Section I of the supplementary material.}

\setlength{\tabcolsep}{4pt}
\begin{table*}[htb]
	\small
	\centering
	\caption{Ablation study of ImpressionGPT on MIMIC-CXR and OpenI datasets. \textbf{Bold} denotes the best result.}
	\label{table1}
        \scalebox{0.9}{
	\begin{tabular}{llp{10mm}<{\centering}p{10mm}<{\centering}p{10mm}<{\centering}p{10mm}<{\centering}p{10mm}<{\centering}p{11mm}<{\centering}}
		\noalign{\smallskip}
		\hline
		\noalign{\smallskip}
		Dataset & Method & R-1$\uparrow$ & R-2$\uparrow$ & R-L$\uparrow$ & {BS-P$\uparrow$} & {BS-R$\uparrow$} & {BS-F1$\uparrow$} \\
		\toprule
		\multirow{4}{*}{MIMIC-CXR~\cite{johnson2019mimic}} &Fixed Prompt
		&33.29    &16.48   &28.35  &86.07  &88.15 &87.07 \\
		& Dynamic Prompt 
		&45.14  &25.75    &38.71  &87.25  &90.04  &88.60  \\
		&Fixed Prompt $+$ Iterative Opt 
		&35.68  &20.30    &31.69  &88.73  &88.46 &88.55 \\ 
		&Dynamic Prompt $+$ Iterative Opt  
		&\textbf{54.45}    &\textbf{34.50}    &\textbf{47.93} &\textbf{91.56}   &\textbf{91.33}  &\textbf{91.41}   \\
		\toprule
		\multirow{4}{*}{OpenI~\cite{demner2016preparing}} &Fixed Prompt
		&33.48    &19.80   &31.33  &87.42  &90.61 &88.94 \\
		& Dynamic Prompt 
		&42.07  &27.02    &39.78  &88.92  &91.76 &90.27  \\
		&Fixed Prompt $+$ Iterative Opt 
		&46.63  &33.17    &44.74  &89.28  &92.28 &90.71 \\ 
		&Dynamic Prompt $+$ Iterative Opt  
		&\textbf{66.37}    &\textbf{54.93}    &\textbf{65.47} &\textbf{93.59}   &\textbf{94.15}  &\textbf{93.83}   \\
		\bottomrule
	\end{tabular}}
\end{table*}
\setlength{\tabcolsep}{1.4pt}

\subsection{{Main Property Studies}}

In this section, we ablate our ImpressionGPT in two aspects: the effectiveness of dynamic prompt and iterative optimization. {In addition to employing Rouge metrics, we introduced BERTScore~\cite{zhang2019bertscore} to further evaluate these two modules. As shown in the Table~\ref{table1}, the upper section displays the results of four methods on the MIMIC-CXR dataset, while the lower section showcases experimental outcomes on OpenI dataset. BS-P, BS-R, and BS-F1 respectively represent the Precision, Recall, and F1 metrics of BERTScore.} As shown in the rows of ``Fixed Prompt", we construct the prompt using fixed examples, which means that for each report in the test set, the prefix instruction in the fixed prompt is consistent. In the rows of {``Dynamic Prompt"} in Table~\ref{table1}, we first search the corpus for examples similar to the test report, and then construct the dynamic prompt. {It can be seen that the experimental results for Rouge-1, Rouge-2, and Rouge-L are improved by \%11.85, \%9.27, and \%10.36 on MIMIC-CXR dataset, and by \%8.59, \%7.22, and \%8.45 on OpenI dataset. {And there is also a noticeable improvement in the BERTScore metrics.} In the rows of ``Fixed Prompt $+$ Iterative Opt", we continue to introduce our iterative optimization algorithm on the fixed prompt and find that the results are further improved on both datasets. And for the OpenI dataset, the results of ``Fixed Prompt $+$ Iterative Opt" even outperform ``Dynamic Prompt". We believe this is due to the fact that OpenI's corpus is relatively small and cannot provide a richer and more similar examples as context, whereas the results after using the iterative optimization method are significantly improved, and the improvement is larger than that on MIMIC-CXR, which has a much larger corpus. This further illustrates the advantages of our iterative optimization algorithm.} In the end, we combine dynamic prompt and iterative optimization, our ImpressionGPT, to achieve the optimal result. The above results show that ChatGPT learns more relevant prior knowledge through the dynamic context we designed, which significantly improves the quality of the generated results. {This demonstrates the feasibility of using a small number of samples to optimize the response of LLMs instead of training the model parameters.} With the introduction of iterative optimization, the model can learn how to generate impression correctly with good responses, while learning how to avoid similar writing styles with bad responses. The model completes self-iterative updates in an interactive way, thus further optimizing the generated results. We also present a case study of our ablation experiments in Section II of supplementary materials.

\setlength{\tabcolsep}{4pt}
\begin{table*}[!t]
\small
\centering
\caption{Comparison results of impressionGPT with other prominent methods on MIMIC-CXR and OpenI dataset. \textbf{Bold} denotes the best result and \underline{Underline} denotes the second-best result.}
\label{table2}
\scalebox{0.9}{
\begin{tabular}{lp{8mm}<{\centering}p{8mm}<{\centering}p{8mm}<{\centering}p{9mm}<{\centering}p{9mm}<{\centering}p{10mm}<{\centering}p{8mm}<{\centering}p{8mm}<{\centering}p{8mm}<{\centering}}
\noalign{\smallskip}
\hline
\noalign{\smallskip}
\multirow{2}{*}{Method} &\multicolumn{6}{c}{MIMIC-CXR~\cite{johnson2019mimic}} &\multicolumn{3}{c}{OpenI~\cite{demner2016preparing}}\\  \cmidrule(r){2-7} \cmidrule(r){8-10}
 & R-1$\uparrow$ & R-2$\uparrow$ & R-L$\uparrow$ & {FC-P$\uparrow$} & {FC-R$\uparrow$} & {FC-F1$\uparrow$}  & R-1$\uparrow$ & R-2$\uparrow$ & R-L$\uparrow$\\
\toprule
LexRank~\cite{erkan2004lexrank}
&18.11    &7.47    &16.87  &-  &-  &-     &14.63    &4.42    &14.06\\
\noalign{\smallskip}
\hline
\noalign{\smallskip}
PGN~\cite{see2017get}   
&46.41    &32.33    &44.76   &54.72  &45.37  &49.61   &63.71    &54.23    &63.38 \\
CGU~\cite{lin2018global}   
&46.50 &32.61 &44.98   &-  &-  &-   &61.60 &53.00 &61.58 \\
\noalign{\smallskip}
\hline
\noalign{\smallskip}
TransEXT~\cite{liu2019text}  
   &31.00    &16.55    &27.49   &-  &-  &-   &15.58    &5.28    &14.42    \\
TransAbs~\cite{liu2019text}
   &47.16    &32.31    &45.47  &56.18  &49.08  &52.39   &59.66    &49.41    &59.18    \\
CAVC~\cite{song2020controlling}
   &43.97 &29.36 &42.50    &-  &-  &-      &53.18 &39.59 &52.86 \\
WGSum (LSTM)~\cite{hu2021word}
   &{47.48}  &{33.03}  &{45.43}  &55.82  &47.13  &51.11    &{64.32}  &\underline{55.48}  &{63.97}   \\
WGSum (Trans)~\cite{hu2021word}
   &{48.37}  &{33.34}  &{46.68}   &56.83  &51.22  &53.88     &{61.63}  &{50.98}  &{61.73}  \\
Jinpeng et al.~\cite{hu2022graph}
   &\underline{49.13}  &\underline{33.76}  &\underline{47.12}  &\underline{58.85}  &\underline{52.33}  &\underline{54.52}   &\underline{64.97}  &\textbf{55.59}  &\underline{64.45}  \\
% ChestXrayBERT~\cite{cai2021chestxraybert}
%    &{41.3*}  &{28.6*}  &{41.5*}  &-  &-  &-  \\
\noalign{\smallskip}
\hline
\noalign{\smallskip}
{ChatGPT}~\cite{openaiIntroducingChatGPT}
&{20.48}  &{9.96}  &{17.02}  &42.83  &57.44  &45.23   &{12.03}  &{3.70}  &{10.52}  \\
{GPT-4}~\cite{openai2023gpt4}
&{19.95}  &{8.58}  &{15.75}  &41.90  &58.04  &44.06  &{11.71}  &{3.43}  &{9.75}    \\
{Radiology-Llama2}~\cite{liu2023radiologyllama2}
&{48.34}  &{32.40}  &{44.27} &43.00  &49.86  &44.44  &{41.85}  &{25.69}  &{40.87}  \\
{Fine-tuned GPT-3}~\cite{brown2020language}
&-  &-  &- &-  &-  &-  &{53.33}  &{41.36}  &{52.39}  \\
\noalign{\smallskip}
\hline
\noalign{\smallskip}
ImpressionGPT (Ours)
  &\textbf{54.45}  &\textbf{34.50}  &\textbf{47.93} &\textbf{79.30}  &\textbf{80.98}  &\textbf{80.09}  &\textbf{66.37}  &54.93  &\textbf{65.47}  \\
\bottomrule
% \multicolumn{7}{l}{\small $*$ denotes the results on a combined dataset of MIMIC-CXR and OpenI.}\\
\end{tabular}}
\end{table*}
\setlength{\tabcolsep}{1.4pt}

\subsection{Comparison with Other Methods}
\label{compare_to_prominent_methods}

Table~\ref{table2} presents a comparison of our ImpressionGPT with other radiology report summarization methods on MIMIC-CXR~\cite{johnson2019mimic} and OpenI~\cite{demner2016preparing} datasets, including graph-based~\cite{erkan2004lexrank}, sequence-based~\cite{see2017get,lin2018global}, pre-trained language model-based methods~\cite{liu2019text,song2020controlling,hu2021word,hu2022graph}, recent GPT-based models~\cite{openai2023gpt4,openaiIntroducingChatGPT}, {and two fine-tuned LLMs, Radiology-Llama2~\cite{liu2023radiologyllama2} and GPT-3~\cite{brown2020language}.} The evaluation is based on the Rouge-1, Rouge-2, Rouge-L, and FC metrics, comparing the impressions generated by our model with handwritten impressions.
% We also include a comparison with ChestXrayBERT~\cite{cai2021chestxraybert}, which used a combined dataset of MIMIC-CXR and OpenI. However, as their data processing method was not public, we use it as a weak reference. 
As shown in Table~\ref{table2}, our method outperforms other methods in all metrics except for the Rouge-2 score on the OpenI dataset, where it performs slightly lower than Jinpeng et al.~\cite{hu2022graph} and WGSum (LSTM)~\cite{hu2021word}. Pre-trained language models perform better than earlier studies, as they are trained with a large amount of medical text data, enabling them to learn the prior knowledge of the medical domain adequately. And our method is even better than the pre-trained language models, with the ability to construct a smaller number of relevant samples in the prompt for the LLM to learn. {In the lower part of Table~\ref{table2}, we compared the original ChatGPT~\cite{openaiIntroducingChatGPT} and GPT-4~\cite{openai2023gpt4} models on MIMIC-CXR and OpenI datasets. It can be seen that the original ChatGPT outperforms the GPT-4 model slightly, indicating that GPT-4 is not as effective as ChatGPT for text summarization tasks. We also compared Radiology-Llama2~\cite{liu2023radiologyllama2}, a large language model fine-tuned on MIMIC-CXR and OpenI datasets. {Recently, the GPT-3 model released a fine-tuning interface, allowing us to utilize the most potent available GPT-3 model, Davinci~\cite{brown2020language}. To save costs, we solely conducted fine-tuning and testing on the relatively small-scale OpenI dataset. Our ImpressionGPT model surpasses Radiology-Llama2, ChatGPT, GPT-4 and the fine-tuned GPT-3.} It's important to note that we achieved this improvement solely through our designed dynamic prompt and iterative optimization framework without training the parameters of the ChatGPT model. Thus, we successfully transfer generic knowledge learned by LLMs in pre-training to domain-specific tasks, such as radiology report summarization, at a lower cost than previous pre-trained language models.}

In summary, we conclude that incorporating semantically similar examples as context in prompt is beneficial in using LLMs in specific domains. Moreover, the generated output of LLMs can be optimized further with iterative interaction.

\section{Discussion and Conclusion}
In this work, we explore the applicability of Large Language Models (LLMs) in the task of radiology report summarization by optimizing the input prompts based on a few existing samples and an iterative scheme. Specifically, relevant examples are extracted from the corpus to create dynamic prompts that facilitate in-context learning of LLMs. Additionally, an iterative optimization method is employed to improve the generated results. The method involves providing automated evaluation feedback to the LLM, along with instructions for good and bad responses. Our approach has demonstrated state-of-the-art results, surpassing existing methods that employ large volumes of medical text data for pre-training. Furthermore, this work is a precursor to the development of other domain-specific language models in the current context of artificial general intelligence \cite{zhao2023brain}.

While developing the iterative scheme of ImpressionGPT, we noticed that evaluating the quality of responses generated by the model is a crucial yet challenging task. In this work, we employed the Rouge-1 score, a conventional metric for calculating text similarity, as the criterion for evaluating the results. We also compared the evaluation criteria using Rouge-1, Rouge-2, and Rouge-L scores and finally found that the performance is sensitive to the set threshold and achieved optimal results using the Rouge-1 score. We speculate that the differences caused by the scores used is due to the fact that the expression of words or phrases in a specific domain differs greatly from the general-domain text used for training the LLMs. Thus, using fine-grained evaluation metrics (i.e., Rouge-1) is better for evaluating the details of the generated results. We also envision that better evaluation criteria that can capture higher-level semantic information from the text will be highly needed with the advancement of LLMs. 

{
The ethical concerns surrounding the clinical application of LLMs are a significant focus~\cite{liu2023summary}.
% ~\cite{liu2023summary,liao2023differentiate,sallam2023chatgpt,lee2023rise,9844014}.
Regarding data privacy, our ImpressionGPT was tested on de-identified data, ensuring patient privacy. Therefore, for clinical applications, employing de-identification algorithms in data pre-processing suffices to safeguard patient privacy without compromising confidentiality.
Regarding model biases, our iterative optimization algorithm automatically constrains the model's generated outputs through automatic evaluations, significantly reducing the possibility of biased or fabricated outputs. Lastly, and most importantly, our model serves to provide valuable reference for clinical experts and does not replace their decision-making authority, thus leaving the final decisions in the hands of healthcare professionals or clinical practitioners.}

In the future, we will continue to optimize the prompt design to better incorporate the domain-specific data from both public and local data sources while at the same time addressing the data privacy and safety concerns involved, especially in a multi-institution scenario. We are also investigating the utilization of knowledge graph in the prompt design to make it more conformed to existing domain knowledge (e.g., the relationship among different diseases). Finally, we will introduce human experts, e.g., radiologists, into the prompt optimization iterations, adding human input to evaluate the generated results when adding them to the prompts. {In such a human-in-the-loop approach, we can better optimize the generated results of LLMs with decisions and opinions from human experts interactively.}

% \section*{Acknowledgments}
% This should be a simple paragraph before the References to thank those individuals and institutions who have supported your work on this article.

\bibliographystyle{IEEEtran.bst}
\bibliography{mybib}


% \end{IEEEbiography}



\vspace{-5pt}
\begin{IEEEbiographynophoto}{Chong Ma} received his master degree of computer science from Northwestern Polytechnical University (Xi’an) in 2019. He is currently pursuing his Ph.D at school of automation of Northwestern Polytechnical University. His main research interests are deep learning, medical image analysis and natural language processing.
\end{IEEEbiographynophoto}

\vspace{-5pt}
\begin{IEEEbiographynophoto}{Zihao Wu} received his B.E. degree from the School of Microelectronics, Tianjin University, Tianjin, China, in 2017 and Master degree in Electrical Engineering and Computer Science Department, Vanderbilt University, Nashville, USA, in 2020. Currently he is pursuing a PhD in computer science at University of Georgia under the supervision of Dr. Tianming Liu. His current research interests include brain inspired AI and deep learning-based medical image analysis.
\end{IEEEbiographynophoto}

\vspace{-5pt}
\begin{IEEEbiographynophoto}{Jiaqi Wang} received her master degree of Statistics from Henu University (Kai’feng) in 2020. She is currently pursuing her Ph.D. at School of Computer Science of Northwestern Polytechnical University. Her main research interests are deep learning, EEG signal analysis and human-computer interaction.
\end{IEEEbiographynophoto}

\vspace{-5pt}
\begin{IEEEbiographynophoto}
{Shaochen Xu} received his B.S. degree in Computer Science from the University of Georgia in 2019. He is currently pursuing his PhD degree at the University of Georgia under the supervision of Dr. Tianming Liu with a research interest in deep learning, natural language processing, and vision transformers.
\end{IEEEbiographynophoto}

\vspace{-5pt}
\begin{IEEEbiographynophoto}{Yaonai Wei} received his master degree of Control science and engineering from Northwestern Polytechnical University (Xi’an) in 2021. He is currently pursuing his Ph.D at school of automation of Northwestern Polytechnical University. His main research interests are deep learning, brain science and artificial general intelligence.
\end{IEEEbiographynophoto}

\vspace{-5pt}
\begin{IEEEbiographynophoto}{Zhengliang Liu} received his B.A. and M.S. degrees in computer science from the University of Wisconsin-Madison and Washington University in St. Louis, in 2018 and 2021, respectively. He is a PhD student in the School of Computing, University of Georgia, Athens, GA. His areas of research include biomedical natural language processing, biomedical image analysis and the intersection of ma-chine learning and radiation oncology.
\end{IEEEbiographynophoto}

\vspace{-5pt}
\begin{IEEEbiographynophoto}{Fang Zeng} is a Data Scientist working at the Massachusetts General Hospital, leading the development of the Large Language Model and multi-agent solutions for medical text processing in healthcare.
\end{IEEEbiographynophoto}

\vspace{-5pt}
\begin{IEEEbiographynophoto}{Prof. Xi Jiang} received the B.E. degree in automation from Northwestern Polytechnical University, Xi’an, China, in 2009, and the Ph.D. degree in computer science from the University of Georgia, Athens, GA, USA, in 2016. Since 2017, he has been an Associate Professor with the School of Life Science and Technology, University of Electronic Science and Technology of China, Chengdu, China. His research interests include machine learning/deep learning-based medical image analysis. Dr. Jiang was a recipient of the Li Foundation Heritage Prize, USA, for “outstanding research and contributions in the interdisciplinary field of brain science” in 2019.
\end{IEEEbiographynophoto}

\vspace{-5pt}
\begin{IEEEbiographynophoto}{Prof. Lei Guo} received the Ph.D. degree from Xidian University, Xi’an, China, in 1994. He is currently a Professor with Northwestern Polytechnical University, Xi’an. His current research interests include computer vision, pattern recognition, and medical image processing.
\end{IEEEbiographynophoto}

\vspace{-5pt}
\begin{IEEEbiographynophoto}{Prof. Xiaoyan Cai} is an associate professor in School of Automation, Northwestern Polytechnical University. She was a research associate in department of computing, the Hong Kong Polytechnic University, Hong Kong, from June 2009 to June 2011. She received the PhD degree from Northwestern Polytechnical University, China, in 2009. Her current research interests include document summarization, information retrieval and machine learning.
\end{IEEEbiographynophoto}

\vspace{-5pt}
\begin{IEEEbiographynophoto}{Prof. Shu Zhang} received the Ph.D. degree in Computer Science from the University of Georgia, USA, in 2018. He is currently a professor at School of Computer Science from the Northwestern Polytechnical University, Xi’an, China. His research interests include biomedical image analysis, brain image analysis, deep learning and machine learning algorithms, artificial intelligence.
\end{IEEEbiographynophoto}

\vspace{-5pt}
\begin{IEEEbiographynophoto}{Prof. Tuo Zhang} received the B.E. and Ph.D. degrees from Northwestern Polytechnical University, Xi’an, China, in 2007 and 2015, respectively. He is currently a Research Associate with Northwestern Polytechnical University. His research interests include machine learning and medical image analysis.
\end{IEEEbiographynophoto}

\vspace{-5pt}
\begin{IEEEbiographynophoto}{Prof. Dajiang Zhu} received the Ph.D. degree from Computer Science at the University of Georgia in U.S. in 2014. His current research interests include machine learning, neuroimaging and computational neuroscience.
\end{IEEEbiographynophoto}

\vspace{-5pt}
\begin{IEEEbiographynophoto}{Prof. Dinggang Shen} (IEEE Fellow, AIMBE Fellow, IAPR Fellow and MICCAI Fellow) is the Founding Dean of the School of Biomedical Engineering at ShanghaiTech University. Before joining ShanghaiTech, he was a tenured Professor of Radiology, Biomedical Research Imaging Center (BRIC), Computer Science, and Biomedical Engineering at the University of North Carolina at Chapel Hill (UNC), USA. Professor Shen has been involved in the application of machine learning and artificial intelligence in medical image computing for a long time, including early brain development, early diagnosis, and the prediction of Alzheimer's disease, as well as diagnosis, prognosis and radiotherapy of brain tumor, breast cancer and prostate cancer. He is a pioneering scientist carrying out imaging AI research all over the world and is one of the first to apply deep learning to medical imaging (2012).
\end{IEEEbiographynophoto}

\vspace{-5pt}
\begin{IEEEbiographynophoto}{Prof. Tianming Liu} is a Distinguished Research Professor of Computer Science at UGA. Dr. Liu's research interests are Brain Imaging, Computational Neuroscience, and Brain-inspired Artificial Intelligence. Dr. Liu has published over 400 papers in these areas, his Google Scholar citations are over 10,000+, and his H-index is 54. Dr. Liu is the recipient of the NIH Career Award and the NSF CAREER Award. Dr. Liu serves on the editorial boards of multiple journals including Medical Image Analysis, IEEE Transactions on Medical Imaging, IEEE Reviews in Biomedical Engineering, IEEE/ACM Transactions on Computational Biology and Bioinformatics, and IEEE Journal of Biomedical and Health Informatics. Dr. Liu is an elected Fellow of the American Institute for Medical and Biological Engineering. 
\end{IEEEbiographynophoto}

\vspace{-5pt}
\begin{IEEEbiographynophoto}{Prof. Xiang Li} is an Assistant Professor at the Massachusetts General Hospital and Harvard Medical School. He received his Ph.D. degree from the Department of Computer Science at the University of Georgia. Dr. Li's research focuses on the Artificial General Intelligence (AGI) and foundation models in healthcare, to tackle the practical challenges of applying AI in a complex clinical context. He has led the development of multiple medical imaging, language processing, EHR analysis, and multi-modal fusion projects, as well as the medical informatics systems for smart data management and AI deployment in the clinical workflow. 
\end{IEEEbiographynophoto}

\newpage
%\appendix

\begin{appendices}
	
\section{Study of Hyper-parameter Selection}
\label{appendix_a}

We ablate our ImpressionGPT in five aspects of preset parameters: the number of examples in fixed or dynamic prompt ($N_s$), the threshold for response judgement in iterative prompt ($T$), the number of good and bad responses in iterative prompt ($Gd$ and $Bd$), and iteration times ($I$). For cost-saving, we select 500 samples in the test set of MIMIC-CXR dataset for our ablation study. This allows us to complete the search for the optimal parameters of the model at a smaller cost. After determining the optimal parameters, we employ the complete test set to get the final results. Table~\ref{teble_ab_fix}$\sim$\ref{teble_ab_iter} show the ablation results of our hyper-parameter study.

\setlength{\tabcolsep}{4pt}
\begin{table}[htb]
	\renewcommand{\arraystretch}{1.2}
	\centering
	\caption{Ablation study of $N_s$ in Fixed and Dynamic Prompt. \textbf{Bold} denotes the best result}
	\label{teble_ab_fix}
	\scalebox{0.9}{
		\begin{tabular}{lccccc}
			\hline
			Type & $N_s$ & R-1$\uparrow$ & R-2$\uparrow$ & R-L$\uparrow$ & Cost \\
			\hline
			\multirow{4}{*}{Fixed} 
			& 5  &29.39    &13.68    &23.74    &\$2  \\
			& 10 &30.21    &13.88    &24.54    &\$3  \\
			& 15 &\textbf{32.57}    &\textbf{15.97}    &\textbf{26.13}    &\$3  \\
			& 18 &32.46    &15.28    &25.81    &\$4  \\ 
			\cline{1-6}
			\multirow{4}{*}{Dynamic} 
			& 5  &41.01    &22.90    &33.45    &\$3  \\
			& 10 &42.53    &24.71    &35.63    &\$4  \\
			& 15 &\textbf{43.97}    &\textbf{24.75}    &\textbf{36.13}    &\$4  \\
			& 18 &43.76    &24.34    &35.63    &\$5  \\ 
			\cline{1-6}
		\end{tabular}
	}
\end{table}
\setlength{\tabcolsep}{1.4pt}

{As shown in the top part of Table~\ref{teble_ab_fix},} we first studied the results of different numbers of examples in fixed prompt, where $N_s$ is the number of examples, and R-1, R-2, R-L, and Cost represent the Rouge-1, Rouge-2, Rouge-L scores and the rough cost, respectively. It can be seen that the best results are obtained by inserting 15 examples in the prompt. When 18 examples are inserted, the results are saturated and the input text may exceed the token limit. 
We continue to investigate the number of examples in the dynamic prompt, {as shown in the bottom part of Table~\ref{teble_ab_fix}.} The experimental results are consistent with {``Fixed Prompt"} that using 15 similar radiology reports as examples in the dynamic prompt performs the best results, and the corresponding Rouge-1, Rouge-2, and Rouge-L scores increase by 11.4\%, 8.78\%, and 10.0\%, respectively, {compared to the ``Fixed Prompt".} This demonstrates that dynamic context constructed using similar examples can help ChatGPT to better perform tasks in specific domain.

\setlength{\tabcolsep}{4pt}
\begin{table}[htb]
	\renewcommand{\arraystretch}{1.2}
	\centering
	\caption{Ablation study of $T$ in Iterative Optimization. \textbf{Bold} denotes the best result}
	\label{teble_ab_thresh}
	\scalebox{0.9}{
		\begin{tabular}{lccccc}
			\hline
			Type & $T$ & R-1$\uparrow$ & R-2$\uparrow$ & R-L$\uparrow$ & Cost \\
			\hline
			\multirow{4}{*}{\begin{tabular}[c]{@{}l@{}}Dynamic\\ $N_s$=15\\ $Gd$=1,$Bd$=n \\ $I$=7\end{tabular}} 
			& 0.50 &50.34    &30.35    &44.06    &\$20  \\
			& 0.65 &51.06    &31.25    &44.16    &\$20  \\
			& 0.70 &\textbf{51.53}    &\textbf{31.95}    &\textbf{44.79}    &\$20  \\
			& 0.75 &51.11    &31.76    &44.33    &\$20  \\ 
			\cline{1-6}
		\end{tabular}
	}
\end{table}
\setlength{\tabcolsep}{1.4pt}

After that, we investigate the value of the metric for evaluating the Impression generated by ChatGPT in iterative optimization, which we noted as $T$, as shown in Table~\ref{teble_ab_thresh}. $N_s$ is the similar examples in dynamic prompt, $Gd$ and $Bd$ represent the number of good and bad examples in iterative prompt, and $I$ is the upper limit of the number of iterations. We find that after using interaction optimization, Rouge-1, Rouge-2, and Rouge-L improve by 7.56\%, 7.2\% and 8.66\%, respectively when $T=0.7$, although the corresponding cost increases about 5 times.

\setlength{\tabcolsep}{4pt}
\begin{table}[htb]
	\renewcommand{\arraystretch}{1.2}
	\centering
	\caption{Ablation study of $Gd$ and $Bd$ in Iterative Optimization. \textbf{Bold} denotes the best result}
	\label{teble_ab_gdbd}
	\scalebox{0.8}{
		\begin{tabular}{lcccccc}
			\hline
			Type & $Gd$ & $Bd$ & R-1$\uparrow$ & R-2$\uparrow$ & R-L$\uparrow$ & Cost \\
			\hline
			\multirow{6}{*}{\begin{tabular}[c]{@{}l@{}}Dynamic\\ $N_s$=15\\ $T$=0.7\\  $I$=7 \end{tabular}} 
			& 0  &1  &50.68   &30.42    &44.08    &\$20  \\
			& 1  &0  &50.31   &29.87    &43.75    &\$20  \\
			& 1  &1  &50.93  &31.15    &44.39    &\$20  \\
			& 1  &n  &\textbf{51.53}    &\textbf{31.95}    &\textbf{44.79}    &\$30   \\ 
			& n  &1  &51.13  &31.05    &44.54    &\$30  \\ 
			& n  &n  &51.46  &31.77    &44.25    &\$40  \\ 
			\hline
		\end{tabular}
	}
\end{table}
\setlength{\tabcolsep}{1.4pt}

\setlength{\tabcolsep}{4pt}
\begin{table}[htb]
	\renewcommand{\arraystretch}{1.2}
	\centering
	\caption{Ablation study of $I$ in Iterative Optimization. \textbf{Bold} denotes the best result}
	\label{teble_ab_iter}
	\scalebox{0.8}{
		\begin{tabular}{llcccc}
			\hline
			Type & $I$ & R-1$\uparrow$ & R-2$\uparrow$ & R-L$\uparrow$ & Cost \\
			\hline
			\multirow{5}{*}{\begin{tabular}[c]{@{}l@{}}Dynamic\\ $N_s$=15\\ $T$=0.7 \\ $Gd$=1,$Bd$=n \end{tabular}} 
			& 7  &{51.53}    &{31.95}    &{44.79}    &\$30  \\
			& 12 &51.91    &32.78    &45.85    & \$50 \\
			& 15 &52.44    &33.13    &46.18    &\$60  \\
			& 17 &\textbf{53.37}    &\textbf{33.68}    &\textbf{46.89}    &\$60  \\
			& 20 &53.07    &33.54    &46.36    &\$70  \\ 
			\cline{1-6}
		\end{tabular}
	}
\end{table}
\setlength{\tabcolsep}{1.4pt}

We continued our study on the number of good and bad examples in the iterative prompt, as shown in Table~\ref{teble_ab_gdbd}, in the $Gd$ and $Bd$ columns, where $0$, $1$ and $n$ represent no examples, one example and multiple examples, respectively. We find that the model achieves optimal results when one good example and multiple bad examples are given. When more good examples are inserted in the prompt, the results are not improved, but rather there are problems with exceeding the textual limit. We believe that with one good example the direction of model optimization can be determined and with multiple bad examples the model generated responses can be further constrained from multiple perspectives. It should be noted that the number of bad examples in the iterative prompt is indefinite and can be appended all the time as long as the input limit is not exceeded.

Finally, we investigate the number of iterations, as shown in Table~\ref{teble_ab_iter}, and find that the model achieves optimal performance when the upper limit number of iterations is set to 17, and its results for Rouge-1, Rouge-2, and Rouge-L improve by 1.84\%, 1.73\%, and 2.1\%, respectively, compared to Table~\ref{teble_ab_gdbd}. Continuing to increase the number of iterations does not bring any performance improvement except for the additional cost. In summary, we determine a set of optimal parameters: ${N_S}=15, T=0.7, {Gd}=1, {Bd}=n, I=17$.

\setlength{\tabcolsep}{4pt}
\begin{table*}[!t]
	\small
	\centering
	\caption{{Examples in our ablation study}}
	\label{table_case_study}
	\scalebox{0.8}{
		\begin{tabular}{lp{180mm}}  
			\noalign{\smallskip}
			\hline
			\noalign{\smallskip}
			\multicolumn{2}{c}{\textbf{Case Information}} \\
			\toprule
			Findings
			&Moderate cardiomegaly is re- demonstrated.  The aorta is tortuous.  Pulmonary vasculature is not engorged.  Patchy opacities are seen in the left lung base, potentially atelectasis but infection or aspiration cannot be excluded.  Streaky atelectasis is also demonstrated in the left lung base.  No pleural effusion or pneumothorax is present.  No acute osseous abnormality is visualized.      \\
			Impression
			&Patchy left basilar opacity may reflect atelectasis, but infection or aspiration cannot be excluded in the correct clinical setting.       \\ 
			\noalign{\smallskip}
			\hline
			\noalign{\smallskip}
			\multicolumn{2}{c}{\textbf{Similar Samples}}   \\
			\noalign{\smallskip}
			\hline
			\noalign{\smallskip}
			No.1  
			& \begin{tabular}[p{180mm}]{@{}p{180mm}@{}}\textbf{Finding:} Heart size is normal. Mediastinal and hilar contours are unremarkable. The pulmonary vasculature is not engorged. Patchy opacities are demonstrated in both lung bases which may reflect atelectasis but infection is not excluded. No pleural effusion or pneumothorax is identified. There are no acute osseous abnormalities. \\\textbf{Iimpression:} Bibasilar patchy opacities may reflect atelectasis but infection is not excluded in the correct clinical setting. \end{tabular}
			\\
			\hline
			No.2  
			& \begin{tabular}[p{180mm}]{@{}p{180mm}@{}}\textbf{Finding:} Cardiac silhouette size is normal.  Mediastinal and hilar contours are within normal limits.  Pulmonary vasculature is normal.  Lungs are hyperinflated.  Patchy opacities noted in the right lung base.  This could reflect atelectasis but infection is not excluded.  Left lung is clear.  No pleural effusion or pneumothorax is demonstrated.  No acute osseous abnormalities seen.  Remote left-sided rib fractures are noted.  Extensive degenerative changes of the left glenohumeral joint are partially imaged. \\\textbf{Iimpression:} Patchy right basilar opacity may reflect atelectasis but infection cannot be excluded in the correct clinical setting. \end{tabular}
			\\
			\hline
			No.3
			& \begin{tabular}[p{180mm}]{@{}p{180mm}@{}}\textbf{Finding:} Cardiac silhouette size is normal.  The mediastinal and hilar contours are unremarkable.  The pulmonary vasculature is not engorged.  Minimal patchy opacity is demonstrated within the right lung base.  No pleural effusion or pneumothorax is present.  No acute osseous abnormality is demonstrated. \\\textbf{Iimpression:} Patchy right basilar opacity may reflect atelectasis however infection or aspiration is not excluded in the correct clinical setting. \end{tabular}
			\\
			\hline
			... &...
			\\
			\hline
			No.14
			& \begin{tabular}[p{180mm}]{@{}p{180mm}@{}}\textbf{Finding:} Lung volumes are low.  The heart size is top normal with a left ventricular predominance.  The mediastinal and hilar contours are unremarkable.  Streaky bibasilar airspace opacities could reflect atelectasis though infection cannot be excluded.  There is no pleural effusion or pneumothorax.  No acute osseous abnormalities are visualized. \\\textbf{Iimpression:} Streaky bibasilar airspace opacities could reflect atelectasis but infection cannot be excluded.  Repeat PA and lateral views with improved inspiratory effort is recommended to better assess the lung bases. \end{tabular}\\
			\noalign{\smallskip}
			\hline
			\noalign{\smallskip}
			\multicolumn{2}{c}{\textbf{Generation Results}}  \\
			\noalign{\smallskip}
			\hline
			\noalign{\smallskip}
			\textbf{Fixed} &\\
			\noalign{\smallskip}
			R-1 = 45.0  &Moderate cardiomegaly. Aorta is tortuous. Patchy opacities in the left lung base suggestive of atelectasis, possibly due to infection or aspiration.\\
			\hline
			\noalign{\smallskip}
			\textbf{Dynamic} &\\
			\noalign{\smallskip}
			R-1 = 57.78  &Patchy opacities in the left lung base, possibly reflecting atelectasis. Infection or aspiration cannot be excluded. Streaky atelectasis is also seen in the left lung base. \\
			\hline
			\noalign{\smallskip}
			\textbf{Fixed + Iteration $17$} &\\
			\noalign{\smallskip}
			R-1 = 52.00  &Cardiomegaly and a tortuous aorta, compatible with underlying cardiovascular disease. Patchy opacities in the left lung base, likely representing atelectasis. Infection or aspiration cannot be excluded. Further evaluation may be required.\\
			\hline
			\noalign{\smallskip}
			\textbf{Dynamic + Iteration} &\\
			\noalign{\smallskip}
			$i=3$ (R-1 = 53.33)
			& Patchy opacities in the left lung base, potentially atelectasis, though infection or aspiration cannot be excluded. Streaky atelectasis is also observed in the left lung base. \\
			$i=5$ (R-1 = 55.00)
			& Moderate cardiomegaly with patchy opacities in the left lung base, potentially representing atelectasis. Infection or aspiration cannot be completely ruled out. \\
			$i=10$ (R-1 = 61.58)
			& Patchy opacities and streaky atelectasis in the left lung base, potentially due to atelectasis. Infection or aspiration cannot be excluded. \\
			$i=15$ (R-1 = 63.83)
			& Moderate cardiomegaly with tortuous aorta. Patchy opacities in the left lung base, potentially atelectasis, with streaky atelectasis. Infection or aspiration cannot be excluded in the correct clinical setting.\\
			$i=17$ (R-1 = 68.42)
			& Moderate cardiomegaly with patchy opacities in the left lung base, likely reflecting atelectasis. Infection or aspiration cannot be excluded. \\
			\bottomrule
		\end{tabular}
	}
\end{table*}
\setlength{\tabcolsep}{1.4pt}

\section{Examples in Our Ablation Study}
\label{appendix_b}

{Table~\ref{table_case_study} presents a test case of our ablation study from Table II of our manuscript. The upper part of Table~\ref{table_case_study} is ``Case Information", which lists the Finding and Impression sections of test report. Our task is to generate an appropriate Impression based on the Finding section. The middle part of Table~\ref{table_case_study} is ``Similar Samples", which are similar examples we constructed in dynamic prompt. These 14 examples are selected from the corpus based on their similarity distance to the Finding section of the test report. And these similar samples (from No.1 to No.14) are ordered from closest to furthest similarity distance to finding section of test report. 
	% The lower part of Table~\ref{table_case_study}, titled ``Iterative Generation", lists the generated results during iterations 3, 5, 10, 15, and 17 along with the corresponding Rouge-1 scores. It can be observed that our similarity search method yields examples highly similar to the test reports, providing customized context information for ChatGPT. Furthermore, with the application of our iterative optimization algorithm, the responses from ChatGPT improve progressively with each iteration.
	The lower of Table~\ref{table_case_study}, titled ``Generation Results", presents the text results and corresponding Rouge-1 scores generated by four different methods in our ablation experiments. ``Fixed" represents the output results obtained using a fixed prompt. It can be observed that the scores are relatively lower, as the contextual information provided by the fixed examples may not be closely related to the actual questions (Test Case), thus lacking assistance. ``Fixed + Iteration" denotes the introduction of our iterative optimization algorithm based on the fixed prompt, where we present the results in the 17th iteration. Noticeable improvements can be observed in the generated results. ``Dynamic" denotes the output result obtained using a dynamic prompt. The ``Similar Samples" in the middle of Table~\ref{table_case_study} demonstrates the high similarity between these examples and the Test Case, indicating that the dynamic prompt provides valuable reference information to ChatGPT, thereby further optimizing its generated results. Lastly, ``Dynamic + Iteration" represents the use of the iterative optimization algorithm on the dynamic prompt, which is our proposed ImpressionGPT. We list the results and Rouge-1 scores for iteration numbers 3, 5, 10, 15, and 17. It can be observed that as the iteration increases, the results generated by ChatGPT also improve progressively.}

\section{{Ablation Study on Corpus Size}}
\label{appendix_corpus_size} 

{In the similarity search module of our model, corpus size serves as a critical factor. Therefore, we contrasted three sets of different-sized corpus using the MIMIC-CXR dataset as an example. As illustrated in Table~\ref{table_corpus_size}, the first row represents ImpressionGPT's similarity search within a corpus of 10,000 samples, the second row presents the experiment with a corpus expanded to 20,000, and the third row indicates the results obtained using the entire training set as the corpus. The experimental outcomes reveal a gradual enhancement in model performance with increasing corpus size. This trend aligns with expectations, as within ImpressionGPT's similarity search module, employing larger corpora yields instances with higher semantic similarity, providing more valuable references for the model.}

\setlength{\tabcolsep}{4pt}
\begin{table}[H]
	\small
	\centering
	\caption{Comparison results of using different size of corpus in similarity calculation. \textbf{Bold} denotes the best result.}
	\label{table_corpus_size}
	\scalebox{0.9}{\begin{tabular}{lccc}
			\noalign{\smallskip}
			\hline
			\noalign{\smallskip}
			\multirow{2}{*}{Corpus Size} &\multicolumn{3}{c}{MIMIC-CXR} \\ \cmidrule(r){2-4} 
			& R-1$\uparrow$ & R-2$\uparrow$ & R-L$\uparrow$  \\
			\hline
			\noalign{\smallskip}
			100,00 Samples
			&{48.32}  &{29.21}  &{41.85}    \\
			200,00 Samples
			&{49.69}  &{30.86}  &{43.16}    \\
			122,014 (Full set)
			&\textbf{54.45}  &\textbf{34.50}  &\textbf{47.93}       \\
			\hline
		\end{tabular}
	}
\end{table}
\setlength{\tabcolsep}{1.4pt}

\end{appendices}

\vfill
\end{CJK}
\end{document}